\documentclass{article} %
\usepackage{iclr2019_conference,times}
\usepackage{enumerate}

\usepackage{amsmath,amsfonts,bm}

\def\eqref#1{equation~\ref{#1}}

\def\1{\bm{1}}

\DeclareMathAlphabet{\mathsfit}{\encodingdefault}{\sfdefault}{m}{sl}
\SetMathAlphabet{\mathsfit}{bold}{\encodingdefault}{\sfdefault}{bx}{n}

\DeclareMathOperator{\sign}{sign}

\PassOptionsToPackage{usenames, dvipsnames}{color}
\usepackage{hyperref}
\usepackage{url}
\usepackage{soul}
\usepackage{graphicx}
\usepackage{comment}
\usepackage{floatrow}
\usepackage{mathtools}
\usepackage{amsthm}
\usepackage{booktabs}
\usepackage{amsmath,amssymb}
\usepackage{amsfonts}
\usepackage{booktabs}
\usepackage{caption}

\usepackage{subcaption}
\usepackage{siunitx}
\usepackage[textsize=tiny,textwidth=8em,backgroundcolor=yellow]{todonotes} 
	
\usepackage{wrapfig}
\usepackage{colortbl}
\definecolor{Lightgray}{RGB}{235,235,235}

\usepackage{arydshln}
\usepackage{wrapfig}
\usepackage{tabularx}

\DeclarePairedDelimiterX{\infdivx}[2]{(}{)}{%
  #1\;\delimsize\|\;#2%
}

\newtheorem{theorem}{Theorem}

\newtheorem{definition}[theorem]{Definition}

\definecolor{c1}{RGB}{239,71,111}
\definecolor{c2}{RGB}{250,131,52}

\title{Exploiting Excessive Invariance caused by Norm-Bounded Adversarial Robustness}

\author{J\"orn-Henrik Jacobsen\thanks{Correspondence to  \it{j.jacobsen@vectorinstitute.ai}} \\ 
Vector Institute and University of Toronto
\And
Jens Behrmannn\\
University of Bremen
\AND
Nicholas Carlini\\
Google Brain
\And
Florian Tram\`er\\
Stanford University
\And
Nicolas Papernot\\
Google Brain
}

\iclrfinalcopy %

\begin{document}

\maketitle

\begin{abstract}

Adversarial examples are malicious inputs crafted to cause a model to misclassify them. Their most common instantiation, ``perturbation-based'' adversarial examples introduce  changes to the input that leave its true label unchanged, yet result in a different model prediction. 
Conversely, ``invariance-based'' adversarial examples insert changes to the input that leave the model's prediction unaffected despite the underlying input's label having changed. 

In this paper, we demonstrate  that robustness to perturbation-based adversarial examples is not only insufficient for general robustness, but worse, it can also increase vulnerability of the model to invariance-based adversarial examples. 
In addition to analytical constructions, we empirically study vision classifiers with state-of-the-art robustness to perturbation-based adversaries constrained by an $\ell_p$ norm. We mount attacks that exploit excessive model invariance in directions relevant to the task, which are able to find adversarial examples \emph{within the $\ell_p$ ball}.
In fact, we find that classifiers trained to be $\ell_p$-norm robust are more vulnerable to invariance-based adversarial examples than their undefended counterparts.

Excessive invariance is not limited to models trained to be robust to perturbation-based $\ell_p$-norm adversaries. In fact, we argue that the term adversarial example is used to capture a series of model limitations, some of which may not have been discovered yet. Accordingly, we call for a set of precise definitions that taxonomize and address each of these shortcomings in learning.

\end{abstract}

\section{Introduction}

Research on adversarial examples is motivated by a spectrum of questions. These range from the security of models deployed in the presence of real-world adversaries to the need to capture limitations of representations and their (in)ability to generalize~\citep{GilmerMotivating2018}. 
The broadest accepted definition of an adversarial example is ``an input to a ML model that is intentionally designed by an attacker to fool the model into producing an incorrect output''~\citep{goodfellow2017attacking}.

To enable concrete progress, many definitions of adversarial examples were introduced in the literature since their initial discovery~\citep{szegedy2013intriguing,biggio2013evasion}. 
In a majority of work, adversarial examples are  commonly formalized as adding a perturbation $\delta$ to some test example $x$ to obtain an input $x^*$ that produces an incorrect model outcome.\footnote{Here, an incorrect output either refers to the model returning any class different from the original \textit{source} class of the input, or a specific \textit{target} class chosen by the adversary prior to searching for a perturbation.} We refer to this entire class of malicious inputs as \textit{perturbation-based adversarial examples}. The adversary's capabilities may optionally be constrained by placing a bound on the maximum perturbation $\delta$ added to the original input (e.g., using an $\ell_p$ norm).

Achieving robustness to perturbation-based adversarial examples, in particular when they are constrained using 
$\ell_p$ norms, is often cast as a problem of learning a
model that is uniformly continuous: the defender wishes to prove that for all $\delta>0$ and for some $\varepsilon>0$, all pairs of points $(x, x^*)$ with $\|x-x^*\|\leq \varepsilon$ satisfy $\|G(x)-G(x^*)\| \leq \delta$ (where $G$ denotes the classifier's logits).
Different papers take different approaches to achieving this result, ranging from robust optimization~\citep{madry2017towards} to training models to have Lipschitz constants~\citep{cisse2017parseval} to models which are provably robust to small $\ell_p$ perturbations ~\citep{kolter2017provable,raghunathan2018certified}.

\begin{figure}
\vspace{-5mm}
\centering
\includegraphics[width=1.\linewidth]{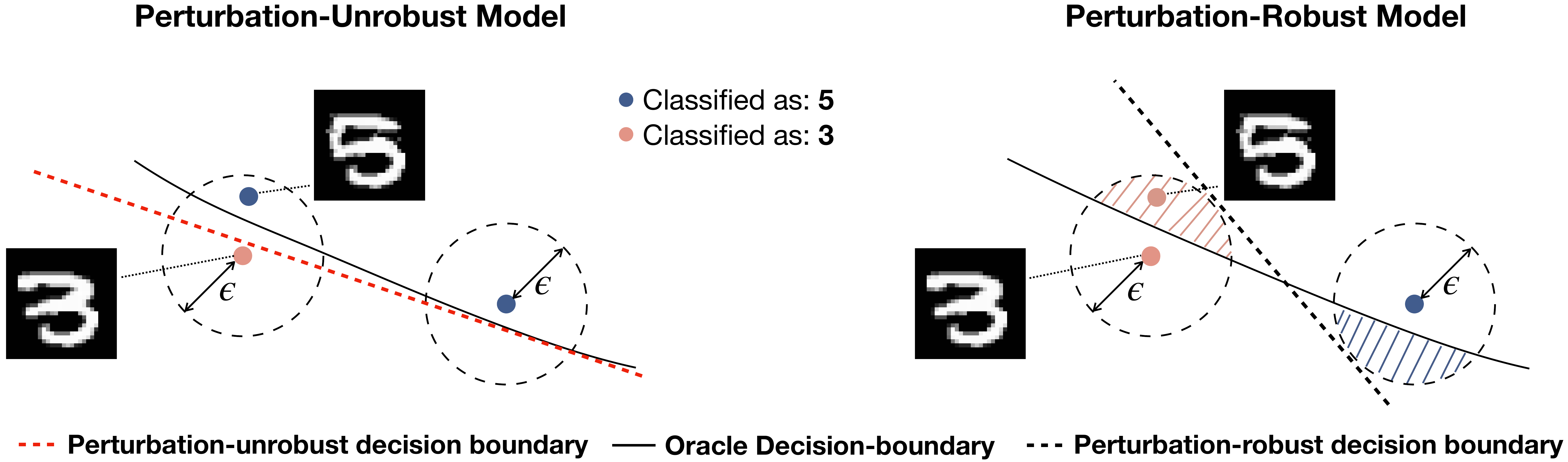}
\caption{[Left]: When training a classifier without constraints, we may end up with a decision boundary that is not robust to perturbation-based adversarial examples. [Right]: However, enforcing robustness to norm-bounded perturbations, introduces erroneous invariance (dashed regions in epsilon spheres). This excessive invariance of the perturbation-robust model in task-relevant directions may be exploited, as shown by the attack proposed in this paper.}
\label{fig:myfig}
\end{figure}

In this paper we present analytical results that show how optimizing for uniform continuity is not only insufficient to address the lack of generalization identified through adversarial examples, but also potentially harmful. Our intuition, captured in Figure~\ref{fig:myfig}, relies on the inability of $\ell_p$-norms to capture the geometry of ideal decision boundaries (or any other distance metric that does not perfectly capture semantics). This leads us to present analytical constructions and empirical evidence that robustness to perturbation-based adversaries can increase the vulnerability of models to other types of adversarial examples.%

Our argument relies on the existence of \textit{invariance-based adversarial examples}~\citep{jacobsen2018excessive}.
Rather than perturbing the input to change the classifier's output, they modify input semantics while keeping the decision of the classifier \textit{identical}. In other words, the vulnerability exploited by invariance-based adversarial examples is a \textit{lack} of sensitivity in directions relevant to the task: the model's consistent prediction does not reflect the change in the input's true label.

Our analytical work exposes a complex relationship between perturbation-based and invariance-based adversarial examples.
We construct a model that is robust to perturbation-based adversarial examples but not to invariance-based adversarial examples. We then demonstrate how an imperfect model for the adversarial spheres task proposed by~\cite{gilmer2018adversarial} is either vulnerable to perturbation-based or invariance-based attacks---depending on whether the point attacked is on the inner or outer sphere. Hence, at least these two  types of adversarial examples are needed to fully account for model failures (more vulnerabilities may be discovered at a later point). 

To demonstrate the practicality of our argument, we then consider vision models with state-of-the-art robustness to $\ell_p$-norm adversaries. We introduce an algorithmic approach for finding invariance-based adversarial examples. Our attacks are model-agnostic and generate $\ell_0$ and $\ell_\infty$ invariance adversarial examples, succeeding at changing the underlying classification (as determined by a human study) in $55\%$ and $21\%$ of cases, respectively. When $\ell_p$-robust models classify the successful attacks, they achieve under $58\%$ (respectively, $5\%$) agreement with the human label.

Perhaps one of the most interesting aspects of our work is to show that different classes of current classifier's limitations fall under the same umbrella term of adversarial examples. Despite this common terminology, each of these limitations may stem from different shortcomings of learning that have non-trivial relationships. To be clear, developing $\ell_p$-norm perturbation-robust classifiers is a useful benchmark task. However, as our paper demonstrates, it is not the only potential way classifiers may make mistakes \emph{even within the $\ell_p$ norm}. Hence, we argue that the community will benefit from working with a series of definitions that precisely  taxonomize  adversarial examples.

\section{Defining Perturbation-based and Invariance-based Adversarial examples}
\label{sec:definitions}

In order to make precise statements about adversarial examples, we begin with two definitions.

\begin{definition}[Perturbation-based Adversarial Examples]
\label{def:advExam}
Let $G$ denote the $i$-th layer, logit or argmax of the classifier. A \textbf{Perturbation-based adversarial example} (or perturbation adversarial) $x^* \in \mathbb{R}^d$ corresponding to a legitimate test input $x \in \mathbb{R}^d$ fulfills:
\begin{enumerate}[(i)]
\item Created by adversary: $x^* \in \mathbb{R}^d$ is created by an algorithm $\mathcal{A}: \mathbb{R}^d \rightarrow \mathbb{R}^d$ with $x \mapsto x^*$.
\item Perturbation of output: $\|G(x^*) - G(x)\| > \delta$ and $\mathcal{O}(x^*) = \mathcal{O}(x)$, where perturbation $\delta > 0$ is set by the adversary and $\mathcal{O}: \mathbb{R}^d \rightarrow \{1, \dots ,C\}$ denotes the \textbf{oracle}.
\end{enumerate}
Furthermore, $x^*$ is \textbf{$\epsilon$-bounded} if $\|x - x^*\| < \epsilon$, where $\|\cdot\|$ is a norm on $\mathbb{R}^d$ and $\epsilon > 0$.
\end{definition}

Property (i) allows us to distinguish perturbation adversarial examples from points that are misclassified by the model without adversarial intervention. Furthermore, the above definition incorporates also adversarial perturbations designed for hidden features as in \citep{sabour2015adversarial}, while usually the decision of the classifier $D$ (argmax-operation on logits) is used as the perturbation target. Our definition also identifies $\epsilon$-bounded perturbation-based adversarial examples~\citep{harnessing_adversarial} as a specific case of unbounded perturbation-based adversarial examples. However, our analysis primarily considers the latter, which correspond to the threat model of a stronger adversary. 

\begin{definition}[Invariance-based Adversarial Examples]
\label{def:advExamInv}
Let $G$ denote the $i$-th layer, logit or argmax of the classifier. An \textbf{invariance-based adversarial example} (or invariance adversarial) $x^* \in \mathbb{R}^d$ corresponding to a legitimate test input $x \in \mathbb{R}^d$ fulfills:
\begin{enumerate}[(i)]
\item Created by adversary: $x^* \in \mathbb{R}^d$ is created by an algorithm $\mathcal{A}: \mathbb{R}^d \rightarrow \mathbb{R}^d$ with $x \mapsto x^*$.
\item Lies in pre-image of $x$ under $G$: $G(x^*) = G(x)$ and $\mathcal{O}(x) \neq \mathcal{O}(x^*)$, where $\mathcal{O}: \mathbb{R}^d \rightarrow \{1, \dots ,C\}$ denotes the \textbf{oracle}.
\end{enumerate}
\end{definition}

As a consequence, $D(x) = D(x^*)$ also holds for invariance-based adversarial examples, where $D$ is the output of the classifier.
Intuitively, adversarial perturbations cause the output of the classifier to change, while the oracle would still label the new input $x^*$ in the original source class. Whereas perturbation-based adversarial examples exploit the classifier's \textit{excessive sensitivity in task-irrelevant directions}, invariance-based adversarial examples explore the classifier's pre-image to identify \textit{excessive invariance in task-relevant directions}: its prediction is unchanged while the oracle's output differs. 
Briefly put, perturbation-based and invariance-based adversarial examples are complementary failure modes of the learned classifier.

\section{Robustness to Perturbation-based Adversarial Examples Can Cause Invariance-based Vulnerabilities}

We now investigate the relationship between the two  adversarial example definitions from Section~\ref{sec:definitions}. So far, it has been unclear whether solving perturbation-based adversarial examples implies solving invariance-based adversarial examples, and vice versa. In the following, we show that this  relationship is intricate and developing models robust in one of the two settings only would be insufficient.

In a general setting, invariance and stability can be uncoupled. For this consider a linear classifier with matrix $A$. The perturbation-robustness is tightly related to forward stability (largest singular value of $A$). On the other hand, the invariance-view relates to the stability of the inverse (smallest singular value of $A$) and to the null-space of $A$. As largest and smallest singular values are uncoupled for general matrices $A$, the relationship between both viewpoints is likely non-trivial in practice.

\subsection{Building our Intuition with Extreme Uniform Continuity}

In the extreme, a classifier achieving perfect uniform continuity would be a constant classifier. 
Let $D: \mathbb{R}^n \rightarrow [0,1]^C$ denote a classifier with $D(x) = y^*$ for all $x\in \mathbb{R}^d$. 
As the classifier maps all inputs to the same output $y^*$, there exist no $x^*$, such that $D(x) \neq D(x^*)$. Thus, the model is trivially perturbation-robust (at the expense of decreased utility).
On the other hand, the pre-image of $y^*$ under $D$ is the entire input space, thus $D$ is arbitrarily vulnerable to invariance-based adversarial examples.
Because this toy model is a constant function over the input domain, no perturbation of an initially correctly classified input can change its prediction.

This trivial model illustrates how one not only needs to control \textit{sensitivity} but also \textit{invariance} alongside \textit{accuracy} to obtain a robust model. Hence, we argue that the often-discussed tradeoff between accuracy and robustness (see~\cite{tsipras2018robustness} for a recent treatment) should in fact take into account at least three notions: accuracy, sensitivity, and invariance. This is depicted in Figure~\ref{fig:myfig}. In the following, we present arguments as for why this insight can also extend to almost perfect classifiers.

\iffalse
While this synthetic sub-optimal model has some symmetry with respect to both perspectives, this is not necessarily the case. For this consider a linear classifier with matrix $A$. The perturbation-robustness is tightly related to forward stability (largest singular value of $A$). On the other hand, the invariance-view relates to the stability of the inverse (smallest singular value of $A$). As largest and smallest singular values are uncoupled for general matrices $A$, the relationship between both viewpoints is most likely non-trivial in practice. 
\fi

\begin{figure}
\vspace{-0.5cm}
\hspace{-.45cm}
\begin{minipage}{0.152\textwidth}
\centering
\includegraphics[width=\textwidth]{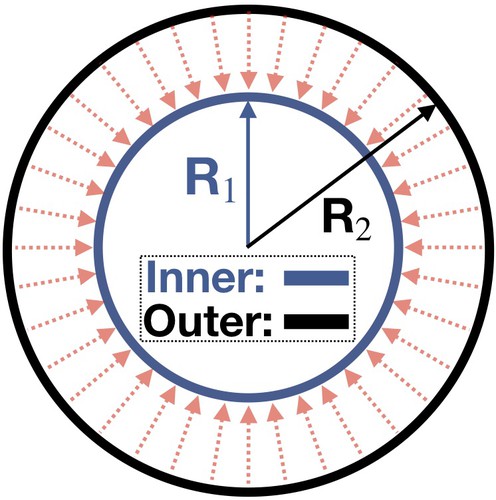}
\end{minipage}
\hspace{-.1cm}
\begin{minipage}{0.37\textwidth}
\centering
\includegraphics[width=\textwidth]{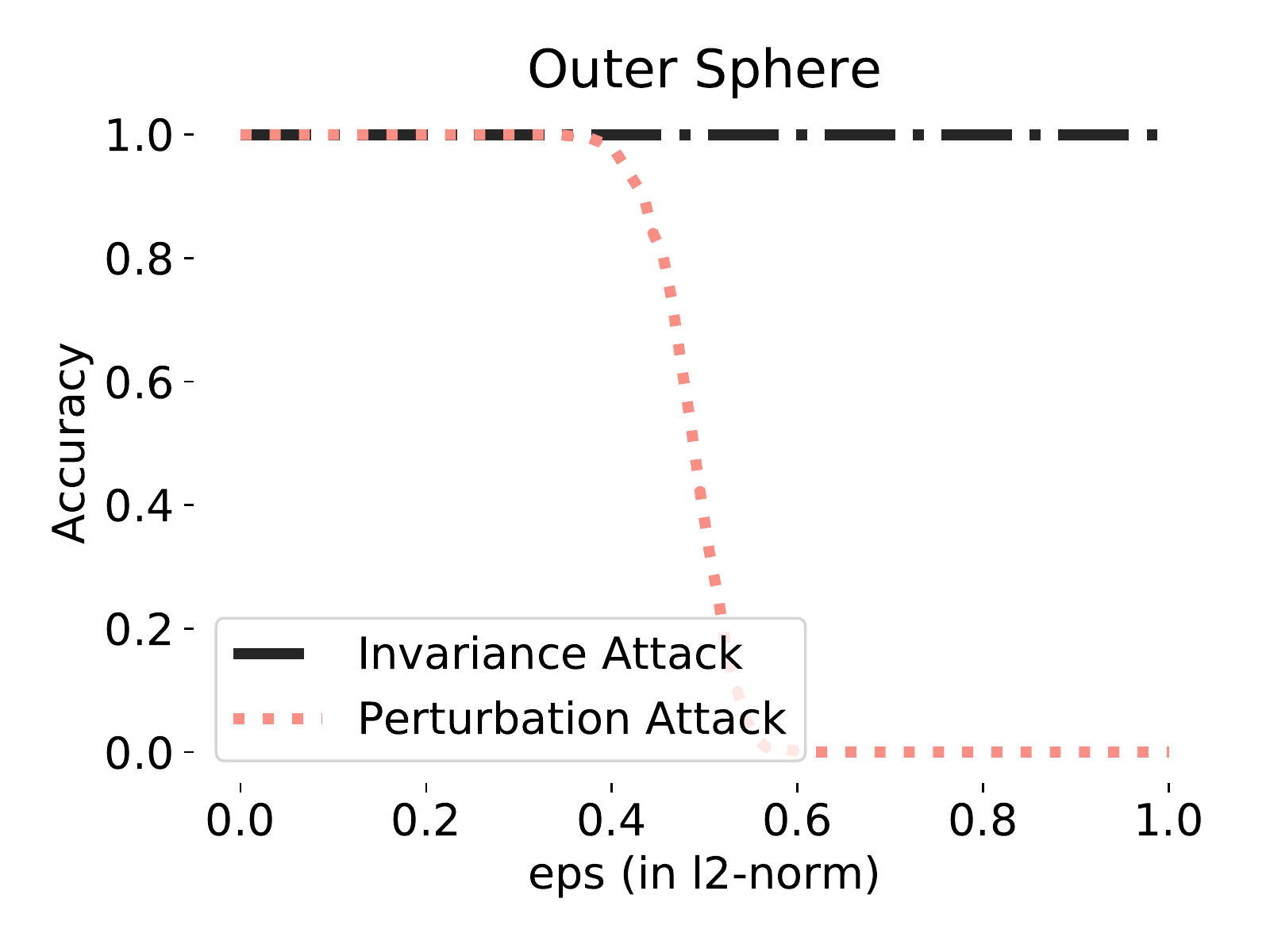}
\end{minipage}
\hspace{-.6cm}
\begin{minipage}{0.15\textwidth}
\centering
\includegraphics[width=\textwidth]{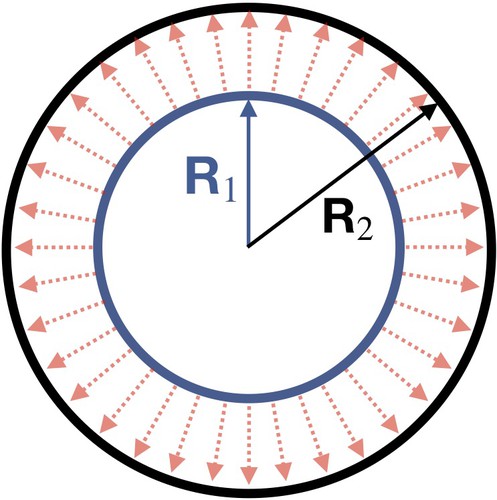}
\end{minipage}
\hspace{-.1cm}
\begin{minipage}{0.37\textwidth}
\centering
\includegraphics[width=\textwidth]{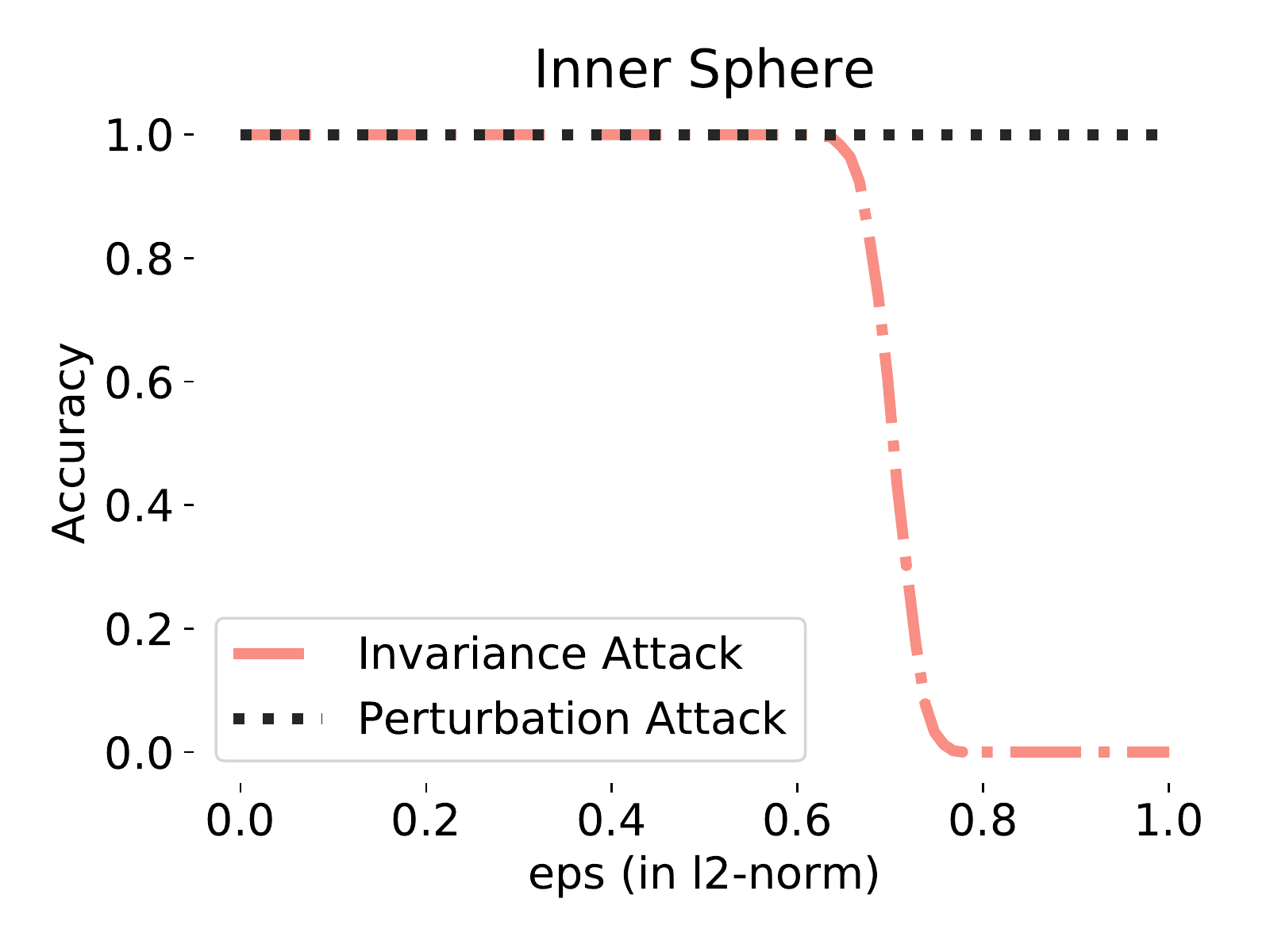}
\end{minipage}
\hspace{-.7cm}
\caption{Robustness experiment on spheres with radii $R_1=1$ and $R_2=1.3$ and max-margin classifier that does not see $n=10$ dimensions of the $d=500$ dimensional input. [Left]: Attacking points from the outer sphere with perturbation-based attacks, with accuracy dropping when increasing the upper bound on $\ell_2$-norm perturbations. [Right]: Attacking points from the inner sphere with invariance-based attacks, with accuracy dropping when increasing the upper bound on $\ell_2$-norm perturbations. Each attack  has a different effect on the manifold. Red arrows indicate the only possible direction of attack for each sphere. Perturbation attacks fail on the inner sphere, while invariance attacks fail on the outer sphere. Hence, both attacks are needed for a full account of model failures.
}
\label{fig:spheres}
\end{figure}

\subsection{Comparing Invariance-based and Perturbation-based Robustness}

We now show how the analysis of perturbation-based and invariance-based adversarial examples can uncover different model failures. To do so, we consider the synthetic \textit{adversarial spheres problem} of~\cite{gilmer2018adversarial}. The goal of this synthetic task is to distinguish points from two cocentric spheres (class 1: $\|x\|_2 = R_1$ and class 2: $\|x\|_2 = R_1$) with different radii $R_1$ and $R_2$. The dataset was designed such that a robust (max-margin) classifier can be formulated as:
\begin{align*}
    D^*(x) = \sign\left(\|x\|_2 - \frac{R_1+R_2}{2}\right).
\end{align*}
Our analysis considers a similar, but slightly sub-optimal classifier in order to study model failures in a controlled setting: 
\begin{align*}
    D(x) = \sign\big(\|x_{1, \dots, d-n}\|_2 - b\big),
\end{align*}
which computes the norm of $x$ from its first $d-n$ cartesian-coordinates and outputs -1 (resp. +1) for the inner (resp. outer) sphere. The bias $b$ is chosen based on finite training set (see Appendix \ref{app:attacksSpheres}).

Even though this sub-optimal classifier reaches nearly 100$\%$ on finite test data, the model is imperfect in the presence of adversaries that operate on the manifold (i.e., produce adversarial examples that remain on one of the two spheres but are misclassified). Most interestingly, the perturbation-based and invariance-based approaches uncover different failures (see Appendix~\ref{app:attacksSpheres} for details on the attacks):
\begin{itemize}
    \item \textbf{Perturbation-based:} All points $x$ from the outer sphere (i.e., $\|x\|_2=R_2$) can be perturbed to $x^*$, where $\mathcal{O}(x) = D(x) \neq D(x^*)$ while staying on the outer sphere (i.e., $\|x^*\|_2=R_2$).
    \item \textbf{Invariance-based:} All points $x$ from the inner sphere ($\|x\|_2 = R_1$) can be perturbed to $x^*$, where $D(x) = D(x^*) \neq \mathcal{O}(x^*)$, despite being in fact on the outer sphere after the perturbation has been added (i.e., $\|x^*\|_2 = R_2$).
\end{itemize}

In Figure~\ref{fig:spheres}, we plot the mean accuracy over points sampled either from the inner or outer sphere, as a function of the norm of the adversarial manipulation added to create perturbation-based and invariance-based adversarial examples. This illustrates  how the robustness regime differs significantly between the two variants of adversarial examples. Therefore, by looking only at perturbation-based (respectively invariance-based) adversarial examples, important model failures may be overlooked. This is exacerbated when the data is sampled in an unbalanced fashion from the two spheres: the inner sphere is robust to perturbation adversarial examples while the outer sphere is robust to invariance adversarial examples (for accurate models).

\section{Invariance-based Attacks in Practice}

We now show that our argument is not limited to the analysis of synthetic tasks, and give practical automated attack algorithms to generate invariance adversarial examples.
We elect to study the only dataset for which robustness is considered to be nearly solved under the $\ell_p$ norm threat model: MNIST~\citep{schott2018towards}. We show that MNIST models trained to be robust to perturbation-based adversarial examples are \emph{less} robust to invariance-based adversarial examples. As a result, we show that while \emph{perturbation} adversarial examples may not exist within the $\ell_p$ ball around test examples, \emph{adversarial examples} still do exist within the $\ell_p$ ball around test examples. 

\paragraph{Why MNIST?} The MNIST dataset is typically a poor choice of dataset for studying adversarial examples, and in particular defenses that are designed to mitigate them~\citep{carlini2019evaluating}. In large part this is due to the fact that MNIST is significantly different from other vision classification problems (e.g., features are quasi-binary and classes are well separated in most cases). %
However, the simplicity of MNIST is why studying $\ell_p$-norm adversarial examples was originally proposed as a toy task to benchmark models~\citep{harnessing_adversarial}. Unexpectedly, it is perhaps much more difficult than was originally expected.
However, several years later, it is now argued that training MNIST classifiers whose decision is constant in an $\ell_p$-norm ball around their training data provides robustness to adversarial examples~\citep{schott2018towards,madry2017towards,kolter2017provable,raghunathan2018certified}.

Furthermore, if defenses relying on the $\ell_p$-norm threat model are going to perform well on a vision task, MNIST is likely the best dataset to measure that---due to the specificities mentioned above. In fact, MNIST is the only dataset for which robustness to adversarial examples is considered even remotely close to being solved~\citep{schott2018towards} and researchers working on (provable) robustness to adversarial examples have moved on to other, larger vision datasets such as CIFAR-10~\citep{madry2017towards,wong2018scaling} or ImageNet~\citep{lecuyer2018certified,cohen2019certified}. 

This section argues that, contrary to popular belief, MNIST is far from being solved. We show why robustness to $\ell_p$-norm perturbation-based adversaries is insufficient, even on MNIST, and why defenses with unreasonably high uniform continuity can harm the performance of the classifier and make it more vulnerable to other attacks exploiting this excessive invariance.

\subsection{A toy worst-case: binarized MNIST classifier}

\begin{wrapfigure}{r}{0.3\textwidth}
\vspace{-12mm}
\includegraphics[width=.7\linewidth]{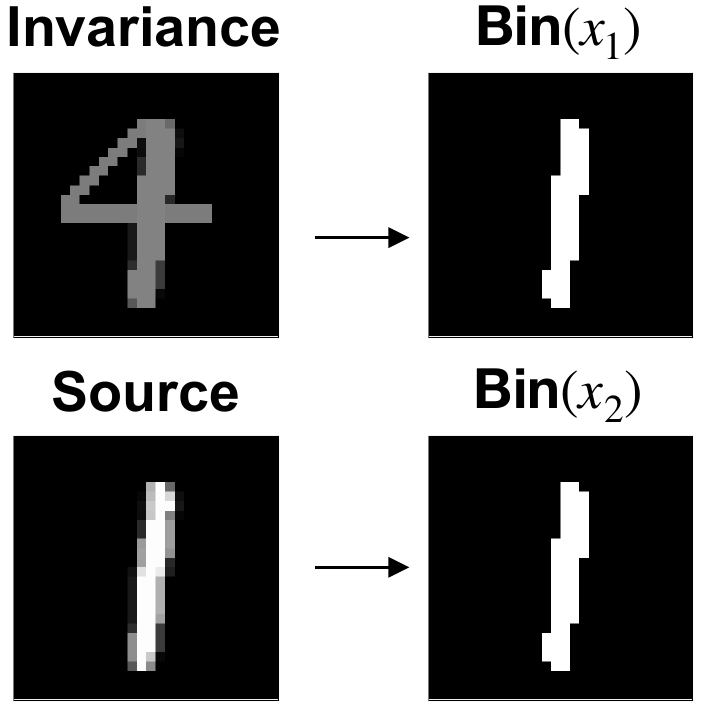}
\caption{Invariance-based adversarial example (top-left) is labeled differently by a human than original (bottom-left). However, both become identical after binarization.\vspace{-3em}}
\label{fig:mnist-invariance-based-adv-x}
\end{wrapfigure}

To give an initial constructive example, consider a MNIST classifier which binarizes (by thresholding at, e.g., 0.5) all of its inputs before classifying them with a neural network. As \citep{tramer2017ensemble,schott2018towards} demonstrate, this binarizing classifier is highly $\ell_\infty$-robust, because most perturbations in the pixel space do not actually change the (thresholded) feature representation.

However, this binary classifier will have trivial invariance-based adversarial examples. Figure~\ref{fig:mnist-invariance-based-adv-x} shows an example of this attack. Two images which are dramatically different to a human (e.g., a digit of a one and a digit of a four) can become identical after pre-processing the images with a thresholding function at $0.5$ (as examined by, e.g., \citet{schott2018towards}).

\subsection{Generating Model-agnostic Invariance-based Adversarial Examples}

In the following, we build on existing invariance-based attacks~\citep{jacobsen2018excessive,jensRelu,li2018study} to propose a model-agnostic algorithm for crafting invariance-based adversarial examples. That is, our attack algorithm generates invariance adversarial examples that cause a human to change their classification, but where most models, not known by the attack algorithm, will \emph{not} change their classification.
Our algorithm for generating invariance-based adversarial examples is simple, albeit tailored to work specifically on datasets where comparing images in pixel space is meaningful, like MNIST.

Begin with a \textit{source} image, correctly classified by both the oracle evaluator (i.e., a human) and the model. Next, try all possible affine transformations of training data points whose label is different from the source image, and find the \textit{target}  training example which---once transformed---has the smallest distance to the source image. Finally, construct an invariance-based adversarial example by perturbing the source image to be ``more similar'' to the target image under the $\ell_p$ metric considered. In Appendix~\ref{app:attacksMNIST}, we describe instantiations of this algorithm for the $\ell_0$ and $\ell_\infty$ norms. Figure \ref{fig:l0attack} visualizes the sub-steps for the $\ell_0$ attack, which are described in details in Appendix~\ref{app:attacksMNIST}.

The underlying assumption of this attack is that small affine transformations are \emph{less likely} to cause an oracle classifier to change its label of the underlying digit than $\ell_p$ perturbations. In practice, we validate this hypothesis with a human study in Section~\ref{sec:eval}. 

\begin{figure}
    \centering
    \includegraphics[scale=.5]{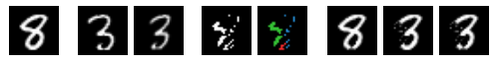} \\
    (a) \hspace{1.8em} (b) \hspace{1.2em} (c) \hspace{1.8em} (d) \hspace{1.2em} (e) \hspace{4.4em} (f-h)  \hspace{2.2em}
    
    \caption{Process for generating $\ell_0$ invariant adversarial examples. From left to right:
    (a) the original image of an 8;
    (b) the nearest training image (labeled as 3), before alignment;
    (c) the nearest training image (still labeled as 3), after alignment;
    (d) the $\delta$ perturbation between the original and aligned training example;
    (e) spectral clustering of the perturbation $\delta$; and
    (f-h) possible invariance adversarial examples, selected by applying subsets of clusters of $\delta$ to the original image. (f) is a failed attempt at an invariance adversarial example. (g) is successful, but introduces a larger perturbation than necessary (adding pixels to the bottom of the 3). (h) is successful and minimally perturbed.}
    \label{fig:l0attack}
\end{figure}

\subsection{Evaluation}
\label{sec:eval}

\paragraph{Attack analysis.} We generate 1000 adversarial examples using each of the two above approaches on examples randomly drawn from the MNIST test set. Our attack is quite slow, with the alignment process taking (amortized) several minutes per example. We performed no optimizations of this process and expect it could be improved.
The mean $\ell_0$ distortion required is 25.9 (with a median of 25). 
The $\ell_\infty$ adversarial examples always use the full budget of $0.3$ and take a similar amount of time to generate; most of the cost is again dominated by finding the nearest test image.

\paragraph{Human Study.} We randomly selected 100 examples from the MNIST test set and create 100 invariance-based adversarial examples under the $\ell_0$ norm and $\ell_\infty$ norm, as described above. We then conduct a human study to evaluate whether or not these invariance adversarial examples indeed are successful, i.e., whether humans agree that the label has been changed despite the model's prediction remaining the same.
We presented 40 human evaluators with these $100$ images, half of which were natural unmodified MNIST digits, and the remaining half were distributed randomly between $\ell_0$ or $\ell_\infty$ invariance adversarial examples. 

\begin{figure}
\begin{subfigure}{.48\textwidth}
    \centering
    \begin{tabular}{l|rr}
    \toprule
         Attack Type & Success Rate & \\
         \midrule
         Clean Images & 0\% \\
         $\ell_0$ Attack & 55\%  \\
         $\ell_\infty$ Attack & 21\%  \\
         \bottomrule
    \end{tabular}
    \caption{Success rate of our invariance adversarial example causing humans to switch their classification.}
    \label{tab:humanstudy:eg}
\end{subfigure}%
\hfill
\begin{subfigure}{.48\textwidth}
\centering
    \includegraphics[scale=.35]{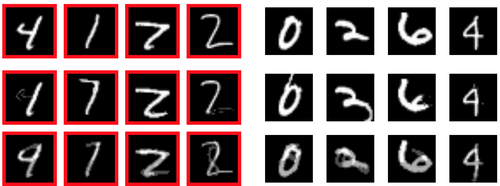}
     \caption{Original test images (top) with our $\ell_0$ (middle) and $\ell_\infty$ (bottom) invariance adversarial examples. \,\,\,\,\,\,\,\, (left) successful attacks; (right) failed attacks.}
    \label{fig:examples}
\end{subfigure}
\caption{Our invariance-based adversarial examples. Humans (acting as the oracle) switch their classification of the image from the original test label to a different label.}
\label{fig:whyislatekborked}
\end{figure}

\paragraph{Results.} For the clean (unmodified) test images, 98 of the 100 examples were labeled correctly by \emph{all} human evaluators. The other 2 images were labeled correctly by over $90\%$ of human evaluators. 

Our $\ell_0$ attack is highly effective: For 48 of the 100 examples at least $70\%$ of human evaluator who saw that digit assigned it the same label, different from the original test label.  Humans only agreed with the original test label (with the same $70\%$ threshold) on 34 of the images, while they did not form a consensus on the remaining 18 examples.
The (much simpler) $\ell_\infty$ attack is less effective: Humans only agreed that the image changed label on 14 of the examples, and agreed the label had not changed in 74 cases. We summarize results in Table~\ref{fig:whyislatekborked} (a).

In Figure~\ref{fig:whyislatekborked} (b) we show sample invariance adversarial examples.
To simplify the analysis in the following section, we split our generated invariance adversarial examples into two sets: the successes and the failures, as determined by whether the plurality decision by humans was different than or equal to the human label.
We only evaluate the models on the subset of invariance adversarial examples that caused the humans to switch their classification.

\paragraph{Model Evaluation.} Now that we have oracle ground-truth labels for each of the images as decided by the humans, we report how often our models agree with the human-assigned label.
Table~\ref{tab:modelaccuracy} summarizes the results of this analysis. For the invariance adversarial examples we report model accuracy only on the \emph{successful} attacks, that is, those where the human oracle label changed between the original image and the modified image.

Every classifiers labeled all successful $\ell_\infty$ adversarial examples \textbf{incorrectly} (with one exception where the $\ell_2$ PGD-trained classifier \cite{madry2017towards} labeled one of the invariance adversarial examples correctly). Despite this fact, PGD adversarial training and Analysis by Synthesis \cite{schott2018towards} are two of the state-of-the-art $\ell_\infty$ perturbation-robust classifiers.

The situation is more complex for the $\ell_0$-invariance adversarial examples. In this setting, the models which achieve \emph{higher} $\ell_0$ perturbation-robustness result in \emph{lower} accuracy on this new invariance test set. For example, \citet{bafna2018thwarting} develops a $\ell_0$ perturbation-robust classifier that relies on the sparse Fourier transform. This perturbation-robust classifier is substantially weaker to invariance adversarial examples, getting only $38\%$ accuracy compared to a baseline classifier's $54\%$ accuracy.

\begin{table}
    \centering
    \begin{tabular}{l p{1.5cm} p{1.5cm} p{1.85cm} p{1.5cm} p{1.5cm} p{1.5cm}}
    \toprule
          & \multicolumn{6}{c}{Fraction of examples where human and model agree} \\
          \midrule
          \textbf{Model:} & \textbf{Baseline} & \textbf{ABS} & \textbf{Binary-ABS} & \textbf{$\ell_\infty$ PGD} & \textbf{$\ell_2$ PGD}  &  \textbf{$\ell_0$ Sparse} \\
         \midrule
         Clean          & 99\% & 99\% & 99\% & 99\% & 99\% & 99\% \\
         $\ell_0$          & 54\% & 58\% & 47\% & 56\%$^*$ & 27\%$^*$ & 38\% \\
         $\ell_\infty$     & 0\% & 0\% & 0\% & 0\% & 5\%$^*$ & 0\%$^*$ \\
         \bottomrule
    \end{tabular}
    \caption{Models which are more robust to \emph{perturbation} adversarial examples (such as those trained with adversarial training) agree with humans \textbf{less often} on \emph{invariance-based} adversarial examples. Agreement between human oracle labels and labels by five models on clean (unmodified) examples and our \emph{successful} $\ell_0$- and $\ell_\infty$-generated invariance adversarial examples. Values denoted with an asterisks $^*$ violate the perturbation threat model of the defense and should not be taken to be attacks. When the model is \emph{wrong}, it classified the input as the original label, and not the new oracle label.}
    \label{tab:modelaccuracy}
\end{table}

\subsection{Natural Images}
\begin{figure}
    \centering
    \includegraphics[scale=.4]{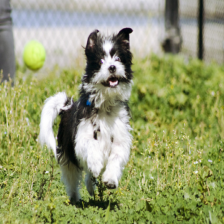}
    \includegraphics[scale=.4]{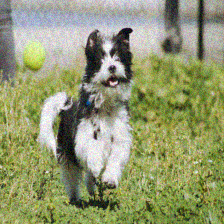}
    \includegraphics[scale=.4]{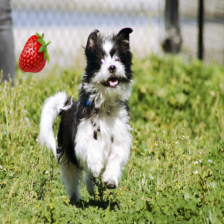} \\
    (a) \hspace{7.5em} (b) \hspace{7.5em} (c)
    \caption{Visualization that large $\ell_2$ norms can also fail to measure semantic changes in images.\,\,\,\, (a) original image in the ImageNet test set labeled as a \emph{tennis ball}; (b) imperceptible perturbation, $\ell_2=24.3$; (c) semantic perturbation with a $\ell_2$ perturbation of $23.2$ that removes the tennis ball.}
    \label{fig:imageNetperturbations}
\end{figure}

While the previous discussion focused on synthetic (Adversarial Spheres) and simple tasks like MNIST, similar phenomena may arise in natural images. In Figure~\ref{fig:imageNetperturbations}, we show two different $\ell_2$ perturbations of the original image (left). The perturbation of the middle image is nearly imperceptible and thus the classifier´s decision should be robust to such changes. On the other hand, the image on the right went through a semantic change (from tennis ball to a strawberry) and thus the classifier should be sensitive to such changes (even though this case is ambiguous due to two objects in the image). However, in terms of the $\ell_2$ norm the change in the right image is even smaller than the imperceptible change in the middle. Hence, making the classifier robust within this $\ell_2$ norm-ball will make the classifier vulnerable to invariance-based adversarial examples like the semantic changes in the right image.

\section{Conclusion}

Training models robust to perturbation-based adversarial examples should not be treated as equivalent to learning models robust to \textit{all} adversarial examples. While most of the research has focused on perturbation-based adversarial examples that exploit excessive classifier \textit{sensitivity}, we show that the reverse viewpoint of excessive classifier \textit{invariance} should also be taken into account when evaluating robustness. Furthermore, other unknown types of adversarial examples may exist: it remains unclear whether or not the union of perturbation and invariance adversarial examples completely captures the full space of evasion attacks.

\paragraph{Consequences for $\ell_p$-norm evaluation.} Our invariance-based attacks are able to find adversarial examples within the $\ell_p$ ball on classifiers that were trained to be robust to $\ell_p$-norm perturbation-based adversaries. 
As a consequence of this analysis, researchers should carefully set the radii of $\ell_p$-balls when measuring robustness to norm-bounded perturbation-based adversarial examples. Furthermore, setting a consistent radius across all of the data may be difficult: we find in our experiments that some class pairs are more easily attacked than others by invariance-based adversaries. 

Some recent defense proposals, which claim extremely high $\ell_0$ and $\ell_\infty$ norm-bounded robustness, are likely over-fitting to peculiarities of MNIST to deliver higher robustness to perturbation-based adversaries. This may not actually be delivering classifiers matching the human oracle more often. 
Indeed, another by-product of our study is to showcase the importance of human studies when the true label of candidate adversarial inputs becomes ambiguous and cannot be inferred algorithmically.

\paragraph{Invariance.} Our work confirms findings reported recently in that it surfaces the need for mitigating undesired invariance in classifiers. The cross-entropy loss as well as architectural elements such as ReLU activation functions have been put forward as possible sources of excessive invariance~\citep{jacobsen2018excessive,jensRelu}. 
However, more work is needed to develop quantitative metrics for invariance-based robustness. One promising architecture class to control invariance-based robustness are invertible networks \citep{dinh2014nice} because,  by construction,  they cannot build up any invariance until the final layer~\citep{jacobsen2018irevnet, behrmann2018invertible}.

\bibliography{iclr2019_conference}
\bibliographystyle{iclr2019_conference}

\appendix

\section{Details about Adversarial Spheres Experiment}
\label{app:attacksSpheres}
In this section, we provide details about the Adversarial Spheres \citep{gilmer2018adversarial} experiment. First, the bias $b$ is chosen, such that the classifier $D$ is the max-margin classifier on the (finite) training set $\mathcal{X}$ (assuming separability:  $l \leq u$): 
\begin{align*}
    l = \max_{\|x\|_2=R_1, x \in T} \|x_{1, \dots, d-n}\|_2, \quad 
    u = \min_{\|x\|_2=R_2, x \in T} \|x_{1, \dots, d-n}\|_2, \quad
    b = l + \frac{u-l}{2}.
\end{align*}
Second, the attacks are designed such that the adversarial examples $x^*$ stay on the data manifold (two concentric spheres). In particular, following steps are taken:

\textbf{Perturbation-based:} All points $x$ from the outer sphere (i.e., $\|x\|_2=R_2$) can be perturbed to $x^*$, where $\mathcal{O}(x) = D(x) \neq D(x^*)$, while staying on the outer sphere (i.e., $\|x^*\|_2=R_2$) via following steps:
\begin{enumerate}
    \item Perturbation of decision: $x^*_{1, \dots, d-n} = a\; (x_{1, \dots, d-n})$, where scaling $a>0$ is chosen such that $\|x^*_{1, \dots, d-n}\|_2 < b$
    \item Projection to outer sphere: $x^*_{d-n, \dots, d} = c\; (x_{d-n, \dots, d})$, where scaling $c>0$ is chosen such that $\|x^*_{d-n, \dots, d}\|_2 = \sqrt{R_2^2 - \|x^*_{1, \dots, d-n}\|_2^2}$
\end{enumerate}
For points $x$ from the inner sphere, this is not possible if $b > R_1$.
    
\textbf{Invariance-based:} All points $x$ from the inner sphere ($\|x\|_2 = R_1$) can be perturbed to $x^*$, where $D(x) = D(x^*) \neq \mathcal{O}(x^*)$, despite being in fact on the outer sphere after the perturbation has been added (i.e., $\|x^*\|_2 = R_2$) via following steps:
\begin{enumerate}
    \item Fixing the used dimensions: $x^*_{1,\dots, d-n} = x_{1,\dots, d-n}$
    \item Perturbation of unused dimensions: $x^*_{d-n, \dots, d} = a\; (x_{d-n, \dots, d})$, where scaling $a>0$ is chosen such that $\|x^*_{d-n, \dots, d}\|_2 = \sqrt{R_2^2 - \|x^*_{1, \dots, d-n}\|_2^2}$
\end{enumerate}

For points $x$ from the outer sphere, this is not possible if $b > R_1$.

\section{Details about Model-agnostic Invariance-based Attacks}
\label{app:attacksMNIST}
Here, we give details about our model-agnostic invariance-based adversarial attacks on MNIST. 

\paragraph{Generating $\ell_0$-invariant adversarial examples.}
Assume we are given a training set $\mathcal{X}$ consisting of labeled example pairs $(x,y)$. As input our algorithm accepts an example $\hat{x}$ with oracle label $\mathcal{O}(\hat{x}) = \hat{y}$. Image $\hat{x}$ with label $\hat{y}=8$ is given in Figure~\ref{fig:l0attack} (a).

Define $\mathcal{S} = \{x : (x,y) \in \mathcal{X}, y \ne \hat{y}\}$, the set of training examples with a different label. Now we define $\mathcal{T}$ to be the set of transformations that we allow: rotations by up to $20$ degrees, horizontal or vertical shifts by up to $6$ pixels (out of 28), shears by up to $20\%$, and re-sizing by up to $50\%$.

Now, we generate the new augmented training set $\mathcal{X}^* = \{(t(x),y,t) : t \in \mathcal{T}, (x,y) \in \mathcal{X}\}$. By assumption, each of these examples is labeled correctly by the oracle. In our experiments, we verify the validity of this assumption through a  human study and omit any candidate adversarial example that violates this assumption. Finally, we search for
\[x^*, y^*, t = \mathop{\text{arg min}}\limits_{(x^*,y^*,t) \in \mathcal{x^*}} \lVert{}x^* - \hat{x}\rVert{}_0. \]
By construction, we know that $\hat{x}$ and $x^*$ are similar in pixel space but have a different label. Figure~\ref{fig:l0attack} (b-c) show this step of the process.
Next, we introduce a number of refinements to make $x^*$ be ``more similar'' to $\hat{x}$. This reduces the $\ell_0$ distortion introduced to create an invariance-based adversarial example---compared to directly returning $x^*$ as the adversarial example. 

First, we define $\delta = |\hat{x}-x^*|>0.5$ where the absolute value and comparison operator are taken element-wise.  Intuitively, $\delta$ represents the pixels that substantially change between $x^*$ and $\hat{x}$. We choose $0.5$ as an arbitrary threshold representing how much a pixel changes before we consider the change ``important''. This step is shown in Figure~\ref{fig:l0attack} (d).
Along with $\delta$ containing the \emph{useful} changes that are responsible for changing the oracle class label of $\hat{x}$, it also contains irrelevant changes that are superficial and do not contribute to changing the oracle class label. For example, in Figure~\ref{fig:l0attack} (d) notice that the green cluster is the only semantically important change; both the red and blue changes are not necessary.

To identify and remove the superficial changes, we perform spectral clustering on $\delta$. We compute $\delta_i$ by enumerating all possible subsets of clusters of pixel regions. This gives us many possible \textbf{potential} adversarial examples $x^*_i = \hat{x}+\delta_i$. Notice these are only potential because we may not actually have applied the necessary change that actually changes the class label. 

We show three of the eight possible candidates in Figure~\ref{fig:l0attack}.
In order to alleviate the need for human inspection of each candidate $x^*_i$ to determine which of these potential adversarial examples is actually misclassified, we follow an approach from Defense-GAN \cite{samangouei2018defense} and the Robust Manifold Defense \cite{ilyas2017robust}: we take the generator from a GAN and use it to assign a likelihood score to the image. We make one small refinement, and use an AC-GAN \cite{mirza2014conditional} and compute the class-conditional likelihood of this image occurring. This process reduces $\ell_0$ distortion by $50\%$ on average.

As a small refinement, we find that initially filtering $\mathcal{X}$ by $20\%$ least-canonical examples makes the attack succeed more often.

\paragraph{Generating $\ell_\infty$-invariant adversarial examples.}
Our approach for generating $\ell_\infty$-invariant examples follows similar ideas as for the $\ell_0$ case, but is conceptually simpler as the perturbation budget can be applied independently for each pixel (as we will see, our $\ell_\infty$ attack is however less effective than the $\ell_0$ one, so further optimizations may prove useful).

We build an augmented training set $\mathcal{X}^*$ as in the $\ell_0$ case. Instead of looking for the closest nearest neighbor for some example $\hat{x}$ with label $\mathcal{O}(\hat{x}) = \hat{y}$, we restrict our search to examples $(x^*, y^*, t) \in \mathcal{X}^*$ with specific target labels $y^*$, which we've empirically found to produce more convincing examples (e.g., we always match digits representing a $1$, with a target digit representing either a $7$ or a $4$). We then simply apply an $\ell_\infty$-bounded perturbation (with $\epsilon=0.3$) to $\hat{x}$ by interpolating with $x^*$, so as to minimize the distance between $\hat{x}$ and the chosen target example $x^*$.

\section{Invariance-based Adversarial Examples for Binarized MNIST}

\begin{figure}[H]
\centering 
\includegraphics[width=.5\linewidth]{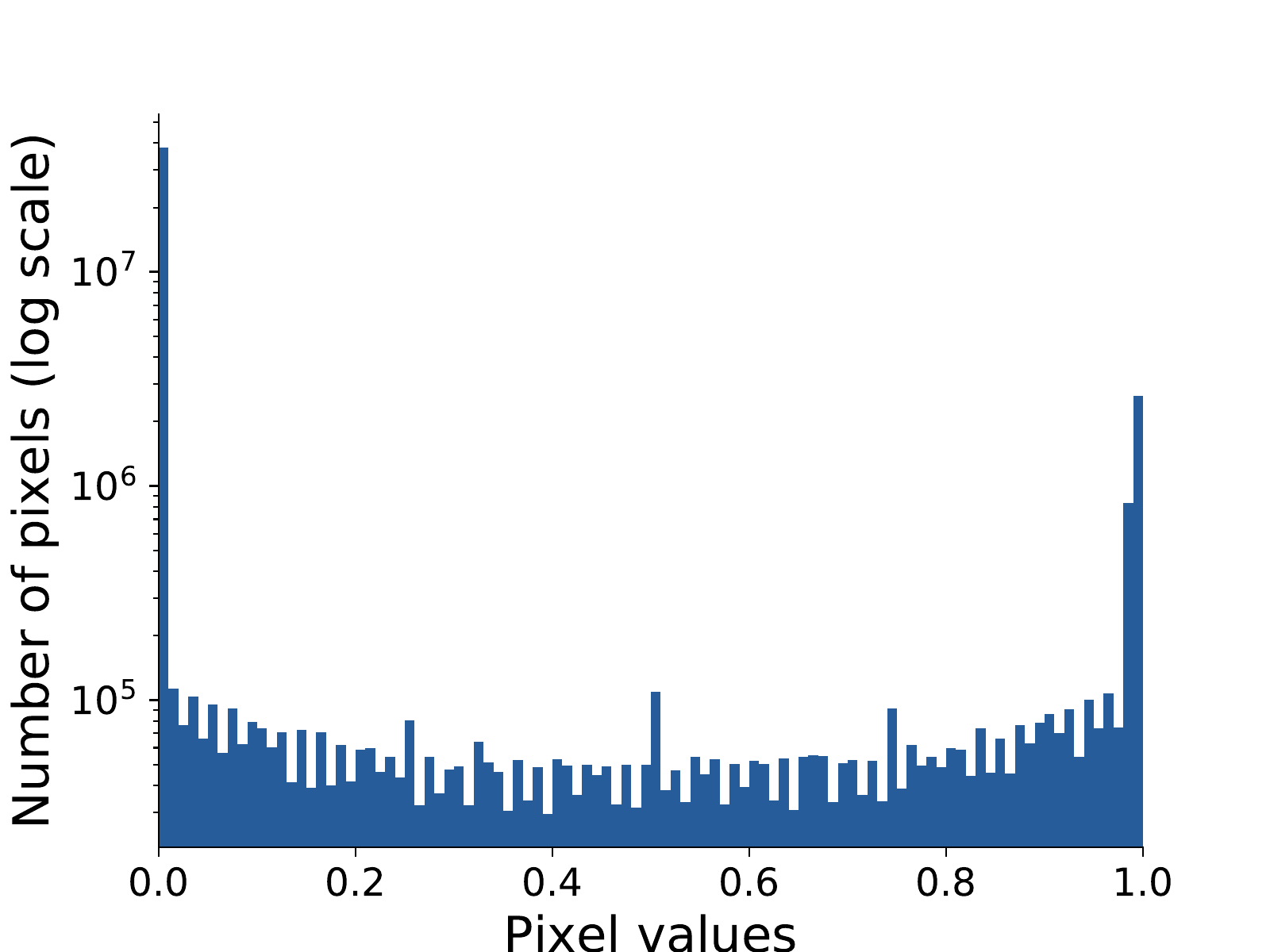}
\caption{Histogram of MNIST pixel values (note the log scale on the y-axis) with two modes around $0$ and $1$. Hence, binarizing inputs to a MNIST model does not impact its performance importantly.}
\label{fig:mnist-pixels-histogram}
\end{figure}

\begin{figure}[H]
\centering
\includegraphics[width=.7\linewidth]{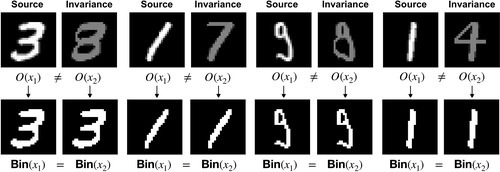}
\caption{Invariance-based adversarial examples for a toy $\ell_\infty$-robust model on MNIST. By thresholding inputs, the model is robust to perturbations $\delta$ such that $\|\delta\|_\infty \lesssim 0.5$. Adversarial examples (top-right of each set of 4 images) are labeled differently by a human. However, they become identical after binarization; the model thus labels both images confidently in the source image's class.}
\label{fig:mnist-invariance-based-adv-x}
\end{figure}

\section{Complete Set of 100 Invariance Adversarial Examples}
Below we give the $100$ randomly-selected test images along with the invariance adversarial examples that were shown during the human study.

\subsection{Original Images}
\includegraphics[scale=.5]{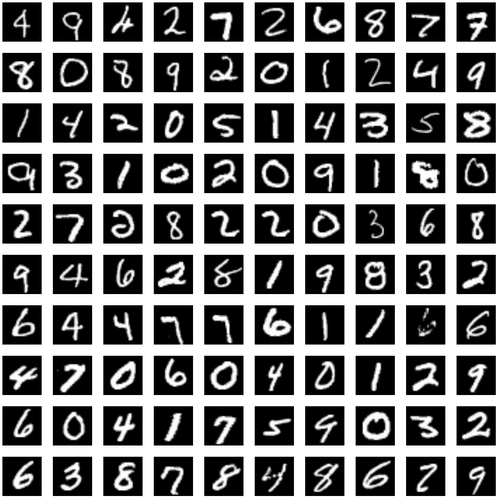}

\subsection{$\ell_0$ Invariance Adversarial Examples}
\includegraphics[scale=.5]{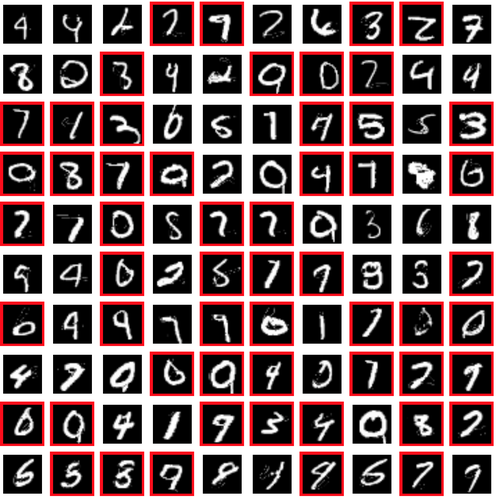}

\subsection{$\ell_\infty$ Invariance Adversarial Examples}
\includegraphics[scale=.5]{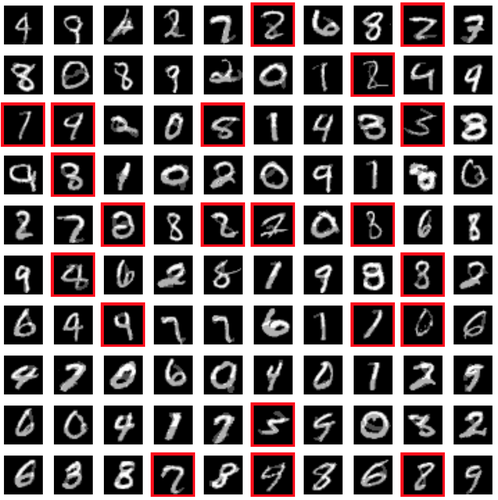}

\end{document}